%% file: root.tex
\documentclass[letterpaper, 10 pt, conference]{class/ieeeconf}
\IEEEoverridecommandlockouts
\usepackage[table]{xcolor}
\usepackage{siunitx}
\usepackage{graphicx}
\usepackage{tikz}

\usepackage{amsmath}
\usepackage{pifont}
\usepackage{tcolorbox}
\usepackage{enumerate}
\let\labelindent\relax
\usepackage[shortlabels]{enumitem}
\usepackage{amssymb}
\usepackage{latexsym}
\usepackage{url}
\usepackage{cite}
\usepackage{relsize}
\usepackage{multirow}
\usepackage{afterpage}
\usepackage{ifthen}
\usepackage{graphicx}
\usepackage{algpseudocode,algorithm,algorithmicx}
\usepackage{subfigure}
\usepackage{booktabs}
\usepackage{makecell}
\usepackage{lipsum}
\usepackage{xcolor} 

\makeatletter
\let\NAT@parse\undefined
\makeatother
\usepackage[bookmarks=false, linkcolor=blue, urlcolor=blue, citecolor=blue]{hyperref} 
 \graphicspath{{./figures/}}

\hypersetup{
    colorlinks=true,
    linkcolor=red,
    filecolor=magenta,      
    urlcolor=blue,
    pdfstartview={FitH},
    citecolor =blue
    }

\definecolor{mygreen}{RGB}{0,150,0}
\newcommand{\cmark}{\textcolor{mygreen}{\ding{51}}}
\newcommand{\xmark}{\textcolor{red}{\ding{55}}}

\title{Adaptive Whole-Body MPC with Optimal Posture Planning for Humanoid Robots under Unknown Payloads
\author{
Manh Cuong Nguyen$^\dagger$, Duc Cuong Vu$^\dagger$ \IEEEmembership{Student Member, IEEE}, Viet Anh Vu, Duc Tuan Pham, \\Toan Nguyen, Minh Hiep Phung, Ngo Anh Vien, Minh Nhat Vu$^*$ \IEEEmembership{Member, IEEE}
\thanks{$^\dagger$ Equal contribution,$^*$ Corresponding author}
\thanks{Manh Cuong Nguyen, Duc Cuong Vu, Viet Anh Vu, Duc Tuan Pham, Toan Nguyen, Minh Hiep Phung and Ngo Anh Vien are with VinRobotics, Vingroup, Hanoi 10000, Vietnam.}
\thanks{Minh Nhat Vu is with Center for AI Research (CAIR), VinUniversity, Hanoi 10000, Vietnam.}
}
}

\newcommand{\rev}[1]{{\color{black}#1}}

\begin{document}

\maketitle

\thispagestyle{empty}
\pagestyle{empty}

\begin{abstract}
Humanoid robots performing lifting and material-handling tasks must maintain balance despite unknown and varying payloads. 
However, payload uncertainty significantly affects whole-body dynamics and may destabilize conventional controllers. 
We propose an adaptive whole-body control architecture that integrates three components: 
(i) a centroidal model predictive controller (MPC) that generates dynamically feasible reference motions, 
(ii) an adaptive fuzzy balance controller that compensates for payload-induced model mismatch by generating corrective torques distributed across the supporting-leg joints, 
and (iii) a waypoint-based optimal posture planner built on a triple-integrator model that produces smooth sit-to-stand and stand-to-sit trajectories satisfying center-of-mass (CoM) stability constraints, joint limits, and whole-body feasibility requirements. 
Together, these components enable stable lowering, grasping, lifting, carrying, and placing behaviors under significant payload uncertainty. 
Extensive simulation and hardware experiments demonstrate that the proposed framework improves balance robustness and reduces tracking degradation during posture transitions and lifting tasks compared to non-adaptive baselines.

\end{abstract}


\input{content/1_intro}
\input{content/2_related_work}

\input{content/3_method}

\input{content/4_experiment}

\input{content/5_discussion}

\bibliographystyle{class/IEEEtran}
\bibliography{class/IEEEabrv,reference}

\end{document}

%% file: content/1_intro.tex
\section{Introduction}
In recent years, humanoid robots have attracted significant attention due to their versatility and potential to operate in human-centered environments \cite{asif2026exploring}. Their comprehensive structure, which integrates mobility \cite{an2025high}, manipulation \cite{liu2025opt2skill}, and perception \cite{bin2025survey}, enables them to perform a wide range of complex tasks and operate in modern manufacturing systems \cite{chi2025towards}. Unlike traditional industrial robots, which are typically designed for structured, repetitive tasks, humanoid robots are inspired by human morphology and motion, enabling them to operate in environments originally designed for human workers. This anthropomorphic configuration provides remarkable advantages in industrial scenarios that require flexibility, dexterity, and whole-body coordination, particularly in tasks involving physical interaction with the environment.

Among these applications, lifting and material handling of unknown payloads represent a critical and challenging problem \cite{purushottam2025heavy}. Such tasks frequently arise in real-world industrial environments where the object properties, including mass, shape, and inertia, are not precisely known in advance. In these situations, the robot must maintain balance while adapting its posture and control strategy to ensure stability and safety. However, despite the promising capabilities of humanoid systems, several technical bottlenecks still limit their practical deployment in industrial lifting tasks. First, the dynamic stability of humanoid robots is highly sensitive to payload variations \cite{mujica2023robust}, as unknown mass distributions can significantly alter system dynamics and shift the CoM, potentially leading to instability or failure. Second, posture transitions during lifting, such as squatting, grasping, and standing up, involve complex whole-body coordination and nonlinear dynamics, which remain difficult to model and control accurately \cite{neunert2018whole}. Third, ensuring robustness and adaptability in the presence of uncertainties, external disturbances, and interaction forces remains a major challenge in humanoid control \cite{liao2023optimization}. Therefore, developing adaptive and robust control frameworks capable of handling unknown payloads, maintaining balance, and optimizing posture during lifting remains a critical research challenge. 

\begin{figure}
    \centering
    \includegraphics[width=\linewidth]{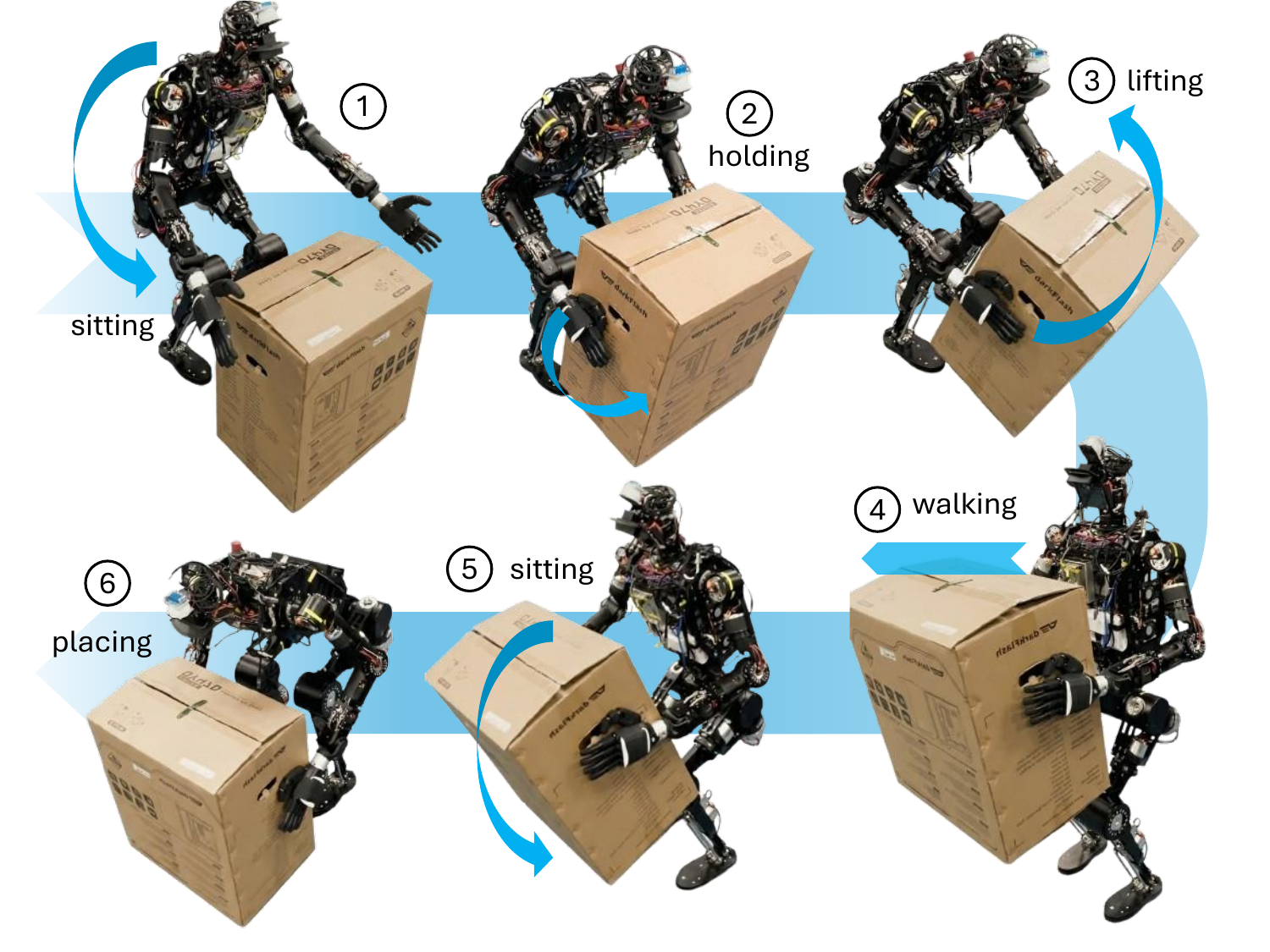}
    \vspace{0.01cm}
    \caption{Illustration of the proposed \textbf{Adaptive Whole-Body MPC with Optimal Posture Planning for Humanoid Robots under of Unknown Payloads}. In stages \textcircled{1}–\textcircled{2}, the robot first lowers its pelvis while regulating pitch. Then, it lifts the unknown payload while tracking the optimal posture transition and performs stable walking \textcircled{3}–\textcircled{4}. Finally, in stages \textcircled{5}–\textcircled{6}, the robot places the object down in a controlled and stable manner. Adaptive fuzzy control ensures stability and constrains CoM deviation throughout the task.}
    \label{fig:punch}
\end{figure}

\begin{table*}
\centering
\caption{Comparison of MPC-based Control Frameworks}
\vspace{4pt}
\renewcommand{\arraystretch}{1.3}
\begin{tabular}{lccccp{5cm}}
\toprule
Methods 
& \makecell{Locomotion} 
& \makecell{Physical\\Constraints} 
& \makecell{Adaptive\\Balancing} 
& \makecell{Posture\\Planning} 
& Application \\
\midrule

Reduced-Order MPC \cite{ghansah2025hierarchical}
& \cmark & \xmark & \xmark & \xmark
& Robust locomotion. \\

MPC + WBC \cite{SleimanWBMPC21}
& \cmark & \cmark & \xmark & \xmark
& Dynamic locomotion, whole-body control. \\

Matrix Exponential-MPC + WBPC \cite{liu2025improved}
& \cmark & \cmark & \xmark & \xmark
& Locomotion.\\

Flexible MPC (FMPC) \cite{yang2025flexible}
& \cmark & \cmark & \xmark & \xmark
&  Variable-speed locomotion. \\

NMPC + Gait Adaptation \cite{li2024adapting}
& \cmark & \cmark & \cmark & \xmark
& Posture-regulating and push-recovery. \\

Kino-dynamic MPC + Gait-Net \cite{li2025gaitnet}
& \cmark & \cmark & \cmark & \xmark
& Variable-frequency locomotion on terrains. \\

Three Particle-MPC + WBC \cite{chen2025squat}
& \xmark & \cmark & \xmark & \cmark
& Continuous and precise whole-body squat. \\

\rowcolor[gray]{0.8}
Adaptive WB-MPC + Posture Planning (Ours)
& \cmark & \cmark & \cmark & \cmark
& Loco-manipulation with unknown payloads. \\

\bottomrule
\end{tabular}
\end{table*}

This paper proposes a comprehensive control architecture that integrates Centroidal MPC with a WBC framework \cite{SleimanWBMPC21}. The centroidal MPC layer generates dynamically feasible CoM motion and contact force references, while the WBC layer ensures accurate joint-space tracking under kinematic and dynamic constraints. To enhance robustness against unknown payload variations, an Adaptive Fuzzy Balance Controller is introduced to generate compensation torques for regulating the CoM and maintaining pelvic stability. The computed compensation torques are subsequently mapped into the joint space of the supporting legs through torque distribution within the whole-body control structure. This mechanism enables adaptive stabilization in the presence of external disturbances and load uncertainties. Furthermore, to ensure smooth and stable execution of sit-to-stand and stand-to-sit motions, a waypoint-based trajectory optimization formulation is proposed. The optimization problem generates dynamically consistent posture transition trajectories while explicitly satisfying CoM stability constraints and joint limits. The resulting motion guarantees stable pelvis balancing and posture transitions that are constraint-consistent.

{Overall, our key contributions of this work are summarized as follows:}
\begin{itemize}
    \item {Firstly, this work presents a novel adaptive control structure that integrates centroidal MPC  with fuzzy logic regulation, ensuring the stability of the overall system under unknown and varying payload conditions during posture and motion tasks.} 
    \item Secondly, an optimization posture transition planning based on their triple-integration waypoint model is provided. This framework helps the robot change posture for the lifting task; the resulting trajectory is smooth, balancing the generated trajectory with joint limitations.
    \item Finally, the simulation and experiment are conducted to validate the performance of the proposed method. 
\end{itemize}

The remainder of this paper is organized as follows. Section~II presents the related work as well as the motivation of this study. Section~III introduces the adaptive balancing control framework, including the robot model, centroidal dynamics, whole-body control, and the proposed method. Section~IV describes the generation of optimal posture-transition motions using the waypoint model. Section~V presents the experimental implementation and results. Finally, Section~VI concludes the paper.

%% file: content/2_related_work.tex
\section{Related works}

\label{sec:related_work}
\subsection{Posture transition for humanoid robot}

Posture transition motions such as sit-to-stand and stand-to-sit are essential capabilities for humanoid robots operating in human-centered environments. These tasks require coordinated whole-body motion, multi-contact interaction, and robustness against model uncertainties and payload variations \cite{gu2025humanoid}. Recent studies have therefore focused on optimization-based approaches to generate dynamically feasible trajectories while explicitly accounting for balance and contact constraints, e.g.,  \cite{neunert2018whole}. Several works have proposed unified centroidal and whole-body MPC formulations to simultaneously optimize contact forces, center-of-mass motion, and kinematic trajectories, enabling robust locomotion and posture \cite{OCS2}. Reduced-order models have also been introduced to improve computational efficiency while maintaining stability, allowing real-time implementation in balancing and posture control tasks \cite{ficht2022direct}. 

Despite these advances, most existing work focuses primarily on locomotion or assumes fixed payload conditions. Efficient and real-time trajectory generation for posture transitions under uncertain payloads remains a challenging problem. In particular, integrating waypoint-based trajectory modeling with MPC to generate smooth, dynamically consistent motions for posture transitions has received limited attention. 

\subsection{Adaptive Fuzzy Control (AFC) for humanoid robot}
Fuzzy logic has been widely applied in humanoid robotics to address nonlinear dynamics and environmental uncertainties. A recent review highlights its effectiveness in balance regulation, gait stabilization, and disturbance rejection when precise models are unavailable \cite{FuzzyHumanoidReview}. 

Fuzzy-based controllers have been successfully used to maintain stability during standing and walking by regulating CoM motion and ensuring the zero moment point containment under uncertainties \cite{Khoi2021FuzzyBiped}. Disturbance rejection performance has been further improved through fuzzy stabilization strategies \cite{Sobirin2021FuzzyStability}, while adaptive fuzzy approaches have enhanced dynamic walking robustness \cite{Chen2019AdaptiveFuzzyPID}. Fuzzy methods have also been extended to terrain adaptation and compliance regulation \cite{Dong2022FuzzyHumanoid}, as well as payload-aware humanoid manipulation \cite{FuzzyHumanoidGrasp2023}. Despite their robustness, conventional fuzzy controllers rely on heuristic rule design and lack predictive optimization and constraint handling. In contrast, MPC enables systematic, constraint-aware torque generation \cite{Ridwan2025SurveyBipedal}, but its performance may degrade under significant uncertainties. 

To combine the strengths of both methods, we integrate Adaptive Fuzzy Control (AFC) with MPC-based Whole-body Control for the pose transition task of humanoids. The adaptive fuzzy component estimates torque compensation from tracking errors and augments the nominal MPC output, improving robustness while preserving the optimal structure of the predictive control framework.

{The choice of a fuzzy compensator over classical model-based alternatives, such as disturbance observers (DOB) \cite{yin2025disturbance}, model-reference adaptive control \cite{vempaty2009model}, or robust control \cite{liang2021fuzzy}, is deliberate and follows from the nature of the payload uncertainty considered here. An unknown payload simultaneously perturbs the robot mass, center-of-mass location, and inertia tensor, so the induced model mismatch is a coupled, state-dependent nonlinearity rather than an additive bounded disturbance. Model-based robust and adaptive laws therefore require either an explicit parametric disturbance model, a reliable uncertainty bound, or online identification of the payload; on a full-scale humanoid these quantities are noisy and slow to converge, and conservative worst-case bounds sacrifice performance. In contrast, the proposed controller is estimation-free: it maps directly measurable balance signals, i.e., the capture-point deviation, the CoM velocity, and its integral, to a bounded compensation moment through an interpretable rule base, without ever estimating the payload mass or reconstructing the disturbance term. This yields three practical advantages. First, it removes the payload-identification loop and its convergence transients, which is critical when the load changes abruptly during grasping and placing. Second, the saturated, rule-bounded output provides an inherent safety envelope on hardware, unlike unbounded adaptive gains that risk parameter drift. Third, the fuzzy law is complementary to, rather than a replacement for, the predictive layer: the MPC retains full constraint-aware optimality for the nominal motion, while the fuzzy term only supplies model-free residual compensation, preserving the MPC structure.}

%% file: content/3_method.tex
\section{Adaptive Balancing for Humanoid Robots}
\label{sec: method}

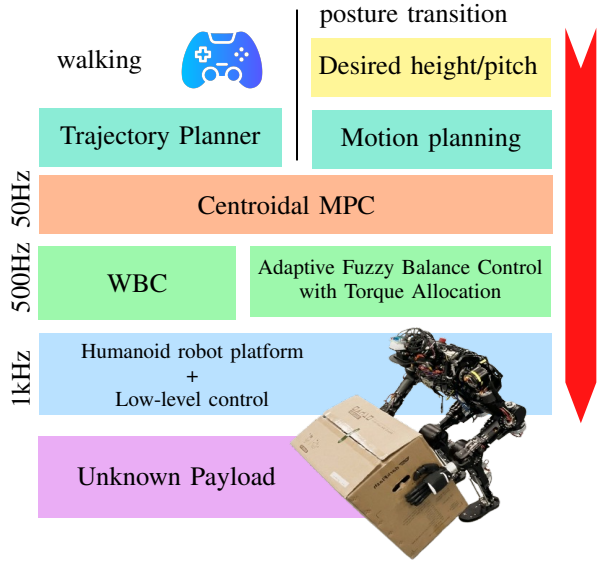
\begin{figure}
    \centering
    \scalebox{1}{\input{tikz/control_structure.tex}}
    \caption{Proposed Adaptive  Whole-Body MPC with Optimal Posture Planning framework.}
    \label{fig:structure}
\end{figure}

This section describes the overall architecture of the proposed fuzzy-based adaptive control framework for the humanoid robot platform. The complete system structure is shown in Fig.~\ref{fig:structure}.

\subsection{Robot Dynamics Modeling}

Let the generalized configuration be $\mathbf{q} = [\mathbf{q}_b^{\top}, \mathbf{q}_j^{\top}]^{\top}$, where $\mathbf{q}_b \in SE(3)$ denotes the floating base and $\mathbf{q}_j \in \mathbb{R}^{n_j}$ represents the actuated joints. The rigid-body dynamics under contact constraints can then be written using the Euler–Lagrange formulation as \cite{Khatib2008}
\begin{subequations}\label{dynamics system}
\begin{align}
&\mathbf{H}_{bb}(\mathbf{q})\ddot{\mathbf{q}}_b + \mathbf{H}_{bj}(\mathbf{q})\ddot{\mathbf{q}}_j + \mathbf{n}_b(\mathbf{q},\dot{\mathbf{q}}) = \mathbf{J}_{cb}^\top(\mathbf{q})\boldsymbol{\lambda}, \label{floating-base dynamics} \\
&\mathbf{H}_{jb}(\mathbf{q})\ddot{\mathbf{q}}_b + \mathbf{H}_{jj}(\mathbf{q})\ddot{\mathbf{q}}_j  + \mathbf{n}_j(\mathbf{q},\dot{\mathbf{q}}) = \boldsymbol{\tau}_j + \mathbf{J}_{cj}^\top(\mathbf{q})\boldsymbol{\lambda}, \label{joint dynamics}\\
&\mathbf{J}_i(\mathbf{q})\ddot{\mathbf{q}} + \dot{\mathbf{J}}_i(\mathbf{q},\dot{\mathbf{q}})\dot{\mathbf{q}} = 0, \quad \forall i \in \mathcal{C},
\end{align}
\end{subequations}
where $\mathbf{H}_{bb}$, $\mathbf{H}_{bj}$, $\mathbf{H}_{jb}$, $\mathbf{H}_{jj}$ denote the mass-inertia matrix and $\mathbf{n}_b(\mathbf{q},\dot{\mathbf{q}})$, $\mathbf{n}_j(\mathbf{q},\dot{\mathbf{q}})$ denote the Coriolis-centrifugal and gravitational effects. The vector $\boldsymbol{\tau}_j$ represents joint actuation torques. Contact interactions are modeled through the stacked \rev{contact Jacobian $\mathbf{J}_c = [\mathbf{J}_{cb},\, \mathbf{J}_{cj}]$, with floating-base part $\mathbf{J}_{cb}$ and joint part $\mathbf{J}_{cj}$,} and corresponding contact wrenches $\boldsymbol{\lambda}$. Finally, the last equation enforces the holonomic constraints for all active contacts $i \in \mathcal{C}$ with contact Jacobian $\mathbf{J}_i(\mathbf{q})$.

\subsection{Centroidal Dynamics and Full Kinematics}

The centroidal dynamics are obtained by applying the Newton–Euler equations at the CoM, which reduces the complexity of the model and enables real-time control performance. Let the centroidal momentum be $\mathbf{h} = [\mathbf{h}_l^{\top}, \mathbf{h}_a^{\top}]^{\top} \in \mathbb{R}^6$, where $\mathbf{h}_l, \mathbf{h}_a \in \mathbb{R}^3$ denote the linear and angular momentum expressed in the centroidal frame. The centroidal dynamics can be represented as
\begin{align}\label{centroidal_dynamics}
\dot{\mathbf{h}} =
\begin{bmatrix}
\sum_{i \in \mathcal{C}}\mathbf{F}_i + m\mathbf{g}\\
\sum_{i \in \mathcal{C}}\big(\mathbf{r}_i \times \mathbf{F}_i + \mathbf{M}_i \big)
\end{bmatrix},
\end{align}
where $\mathbf{r}_i \in \mathbb{R}^3$ is the $i$-th contact position relative to the CoM, 
$\mathbf{F}_i, \mathbf{M}_i \in \mathbb{R}^3$ are the corresponding contact force and moment, $m$ denotes the robot's mass and $\mathbf{g} = [0,0,g]^{\top}$ with $g$ representing the gravity constant.

Using the centroidal momentum matrix $\mathbf{A}(\mathbf{q})$, the momentum is related to the generalized velocities by $\mathbf{h} = \mathbf{A}(\mathbf{q})\dot{\mathbf{q}}$. By defining the system state as $\mathbf{x} = [\mathbf{h}^{\top},\mathbf{q}_b^{\top}, \mathbf{q}_j^{\top}]^{\top}$ and the input vector $\mathbf{u}$ consisting of joint velocities and contact wrenches, the overall planning dynamics can be compactly expressed as
\begin{align}
    \dot{\mathbf{x}} = \mathbf{f}(\mathbf{x},\mathbf{u}) \:.
\end{align}

\subsection{MPC-based Whole-body Control} 
A nonlinear MPC framework over a prediction horizon $T>0$ is proposed as \cite{SleimanWBMPC21}
\begin{subequations}\label{centroidal MPC}
\begin{align}
    \min_{\mathbf{u}(\cdot)} \quad 
    & \Phi(\mathbf{x}(T)) 
    + \int_{0}^{T} l(\mathbf{x}(\rho), \mathbf{u}(\rho))\, d\rho, \\
    \text{s.t.} \quad 
    & \dot{\mathbf{x}} = \mathbf{f}(\mathbf{x},\mathbf{u}), \quad \mathbf{x}(0) = \mathbf{x}_t \\
    &(\mathbf{x}, \mathbf{u}) \in \mathcal{X} \times \mathcal{U}.
\end{align} 
\end{subequations}
where $l(\mathbf{x},\mathbf{u})$ and $\Phi(\mathbf{x}(T))$ denote the stage and terminal costs, respectively, and $\mathcal{X}$ and $\mathcal{U}$ represent the state and input constraint sets. The optimization problem in \eqref{centroidal MPC} ensures that the planned motion satisfies these constraints. \rev{As in \cite{SleimanWBMPC21}, they include friction-cone and center-of-pressure constraints on the stance-foot wrenches, zero stance-foot velocity, zero wrench and a normal-velocity profile for swing feet, and joint position and velocity limits.}
To compute the joint torques with low computational cost, we employ an inverse dynamics approach. After solving \eqref{centroidal MPC}, the first control input is obtained as $\mathbf{u}(0) = [\rev{\dot{\mathbf{q}}_j^{*\top}}, \boldsymbol{\lambda}^{*\top}]^\top$. The desired joint acceleration is chosen to ensure exponential tracking as
$\ddot{\mathbf{q}}_j^* = -K_p \mathbf{e}_j - K_d \dot{\mathbf{e}}_j$,
where $K_p$ and $K_d$ are proportional and derivative gains, $\mathbf{e}_j = \mathbf{q}_j - \mathbf{q}_j^*$ and $\dot{\mathbf{e}}_j = \dot{\mathbf{q}}_j - \rev{\dot{\mathbf{q}}_j^*}$. 
The corresponding floating-base acceleration is computed from \eqref{floating-base dynamics} as
$\ddot{\mathbf{q}}_b^* 
= 
\mathbf{H}_{bb}^{-1}
\left(
\mathbf{J}_{cb}^\top \boldsymbol{\lambda}^*
-
\mathbf{H}_{bj}\ddot{\mathbf{q}}_j^*
-
\mathbf{n}_b
\right).$
Substituting $\ddot{\mathbf{q}}_b^*$ and $\ddot{\mathbf{q}}_j^*$ into the joint dynamics yields the torque command
\begin{align}
\boldsymbol{\tau}_j 
= 
\mathbf{H}_{jb}\ddot{\mathbf{q}}_b^*
+
\mathbf{H}_{jj}\ddot{\mathbf{q}}_j^*
+
\mathbf{n}_j
-
\mathbf{J}_{cj}^\top \boldsymbol{\lambda}^*.
\end{align}
We adopt inverse dynamics instead of a task-space quadratic programming (TS-QP) controller as in \cite{dai2025romocoroboticmotioncontrol} because all dynamic feasibility and contact constraints are already handled at the centroidal Nonlinear MPC level. The lower layer, therefore, only needs to track the planned accelerations and contact forces. \rev{This avoids an additional optimization problem and task-weight tuning.}

\subsection{Adaptive Fuzzy Inference System}

To correct residual tracking errors and external disturbances that may cause the robot to deviate from a stable posture, we add an adaptive fuzzy balance controller that computes a compensation moment, which is then allocated to leg joint torques. The controller aims to reduce the error between the Capture Point and the center of the support polygon. Under the Linear Inverted Pendulum assumption, the Capture Point in the horizontal plane is defined as $\boldsymbol{\xi} = \mathbf{r}_{\mathrm{com}} + \dot{\mathbf{r}}_{\mathrm{com}}/\omega$, where $\mathbf{r}_{\mathrm{com}} \in \mathbb{R}^2$ is the ground projection of the CoM, $\dot{\mathbf{r}}_{\mathrm{com}}$ is the CoM velocity, and \rev{$\omega = \sqrt{g/z_c}$ with $z_c$} the nominal CoM height. Let $\mathbf{c}_{\mathrm{supp}} \in \mathbb{R}^2$ denote the center of the support polygon, computed as the weighted centroid of the stance foot contact points. The balance error is
\begin{align}\label{balance_error}
\mathbf{e} &= \boldsymbol{\xi} - \mathbf{c}_{\mathrm{supp}} = \mathbf{r}_{\mathrm{com}} - \mathbf{c}_{\mathrm{supp}} + \dfrac{\dot{\mathbf{r}}_{\mathrm{com}}}{\omega}.
\end{align}
We express $\mathbf{e}$ in the robot's base frame as \rev{$\mathbf{e} = [e_x, e_y]^\top$. Note that $\mathbf{e}$ is a displacement (in m), not an angle: the forward displacement $e_x$ is corrected by a moment about the pitch axis, and the lateral displacement $e_y$ by a moment about the roll axis.}. The integral of the error $\mathbf{e}_{\mathrm{int}} = \int \mathbf{e} \, dt$ is accumulated with anti-windup clamping to $|e_{\mathrm{int}}| \leq e_{\mathrm{int},\max}$, reducing steady-state error. The fuzzy controller takes three inputs per axis: the deviation $e$, the CoM velocity $\dot{r}$, and the integral $e_{\mathrm{int}}$, and outputs a compensation moment $\boldsymbol{\tau}_{\mathrm{comp}}^{\mathrm{task}} = [\tau_{\mathrm{pitch}}, \tau_{\mathrm{roll}}]^\top$ (in Nm).

The controller implements a four-layer fuzzy inference system with Gaussian membership functions.

\textbf{Layer 1 (Fuzzification).}
Each crisp input is mapped to membership degrees over five linguistic variables: NB (Negative Big), NS (Negative Small), Z (Zero), PS (Positive Small), PB (Positive Big). The Gaussian membership function for set $\ell \in \{\mathrm{NB},\mathrm{NS},\mathrm{Z},\mathrm{PS},\mathrm{PB}\}$ is
\begin{align}\label{gaussian_mf}
\mu_\ell(x) = \exp\left(-\dfrac{(x - c_\ell)^2}{2\sigma_\ell^2}\right),
\end{align}
where $c_\ell$ is the center and $\sigma_\ell$ the width.

\textbf{Layer 2 (Rule base).}
The rule base consists of $5 \times 5 \times 5 = 125$ rules per axis, indexed by $(i,j,k)$ for deviation, velocity, and integral. Each rule has the form
\begin{align}
    & \text{IF } e \text{ is } \mathcal{A}_i \text{, AND } \dot{r}  \text{ is }  \mathcal{B}_j \text{, AND } e_{\mathrm{int}}  \text{ is } \mathcal{C}_k \notag\\
    & \text{THEN consequent } y_{ijk}.
\end{align}
The firing strength of rule $(i,j,k)$ is computed with the product t-norm:
\begin{align}\label{firing_strength}
w_{ijk} = \mu_{\mathrm{dev}}^{(i)}(e) \cdot \mu_{\mathrm{vel}}^{(j)}(\dot{r}) \cdot \mu_{\mathrm{int}}^{(k)}(e_{\mathrm{int}}).
\end{align}

\textbf{Layer 3 (Consequent).}
The consequent $y_{ijk}$ of each rule combines a base term from the deviation--velocity table and an additive integral effect:
\begin{align}\label{consequent}
y_{ijk} = \mathrm{clip}_{[-1,1]}\big(y_{\mathrm{base}}^{(ij)} + \Delta_{\mathrm{int}}^{(k)}\big),
\end{align}
where $y_{\mathrm{base}}^{(ij)}$ encodes the PD-like policy and $\Delta_{\mathrm{int}}^{(k)}$ adds integral action to eliminate steady-state error.

\textbf{Layer 4 (Defuzzification).}
The output is the weighted average of consequents, scaled and saturated:
\begin{align}\label{fuzzy_output}
\tau = S \cdot \dfrac{\sum_{i,j,k} w_{ijk} \, y_{ijk}}{\sum_{i,j,k} w_{ijk}}, \qquad |\tau| \leq \tau_{\max}.
\end{align}
where $S$ is an axis-dependent output scale. The compensation moment $\boldsymbol{\tau}_{\mathrm{comp}}^{\mathrm{task}} = [\tau_{\mathrm{pitch}}, \tau_{\mathrm{roll}}]^\top$ is computed independently for each axis.

\begin{figure}
    \centering
    \includegraphics[width=1.10\linewidth]{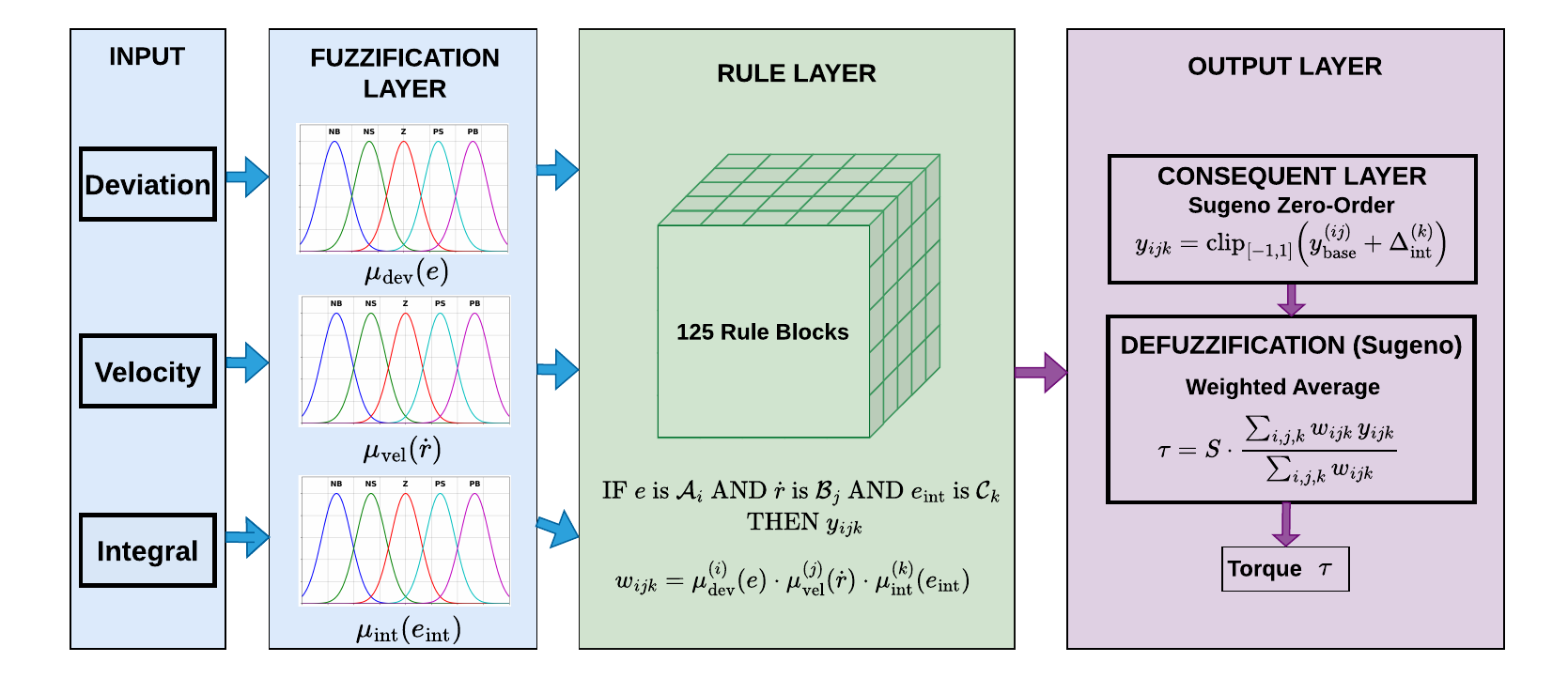}
    \vspace{0.01cm}
    \caption{The Fuzzy Inference System contains four layers for estimating the torque compensation.}
    \label{fig:fuzzy}
\end{figure}


\rev{The rule base is hand-crafted with a symmetric PD-type layout. These parameters $c_\ell$, $\sigma_\ell$, $S$, $\tau_{\max}$ were tuned by expert experience. The same parameters are used in all experiments. The law is closely related to a saturated, gain-scheduled PID on the capture-point error, but its effective gains vary jointly with $e$, $\dot r$, and $e_{\mathrm{int}}$, e.g., a decaying deviation is not over-corrected; a quantitative comparison with PID is left for future work.}

\subsection{Torque allocation}

With the compensation torque $\boldsymbol{\tau}_{\mathrm{comp}}^{\mathrm{task}}$ derived from the adaptive fuzzy framework, a torque allocation scheme is required to distribute this compensation effort to the leg joints. \rev{The compensation is applied only through the legs, because the arms hold the payload and corrective upper-body motion would disturb the grasp.} A commonly used approach, motivated by optimization-based force and torque allocation methods, can be formulated for the stance leg as follows:
\begin{subequations}
\begin{align}
    \boldsymbol{\tau}_{\mathrm{leg}}^*  \quad & = \quad \arg\min \quad \boldsymbol{\tau}_{\mathrm{leg}}^\top\boldsymbol{\tau}_{\mathrm{leg}},\\
    \text{s.t.} \quad& \quad \Omega(\boldsymbol{\tau}_{\mathrm{leg}}) = \boldsymbol{\tau}_{\mathrm{comp}}^{\mathrm{task}}.
\end{align}
\end{subequations}
Here, $\boldsymbol{\tau}_{\mathrm{leg}}^*$ denotes the additional torque applied to the low-level controller to enhance the robustness and stability of the center-of-mass (CoM) motion. The mapping $\Omega(\cdot)$ represents the relationship between the joint torques of the stance leg and the resulting task-space torque at the pelvis (treated as the end-effector) during the stance phase.

\input{content/3_1_posture_transition}

%% file: tikz/control_structure.tex
\tikzset{every picture/.style={line width=0.75pt}} 

\begin{tikzpicture}[x=0.75pt,y=0.75pt,yscale=-1,xscale=1]

\draw  [draw opacity=0][fill={rgb, 255:red, 255; green, 21; blue, 21 }  ,fill opacity=1 ] (552.73,15.33) -- (552.73,193.34) -- (544.95,213.84) -- (537.18,193.34) -- (537.18,15.33) -- (544.95,35.83) -- cycle ;
\draw  [draw opacity=0][fill={rgb, 255:red, 222; green, 123; blue, 240 }  ,fill opacity=0.62 ] (272.47,220.7) -- (439.87,220.7) -- (439.87,261.39) -- (272.47,261.39) -- cycle ;
\draw  [draw opacity=0][fill={rgb, 255:red, 255; green, 164; blue, 115 }  ,fill opacity=0.77 ] (272.97,89.37) -- (530.86,89.37) -- (530.86,118.94) -- (272.97,118.94) -- cycle ;
\draw  [draw opacity=0][fill={rgb, 255:red, 112; green, 245; blue, 134 }  ,fill opacity=0.68 ] (272.97,124.94) -- (371.87,124.94) -- (371.87,161.79) -- (272.97,161.79) -- cycle ;
\draw  [draw opacity=0][fill={rgb, 255:red, 112; green, 245; blue, 134 }  ,fill opacity=0.68 ] (378.86,124.94) -- (529.86,124.94) -- (529.86,160.29) -- (378.86,160.29) -- cycle ;
\draw  [draw opacity=0][fill={rgb, 255:red, 80; green, 227; blue, 194 }  ,fill opacity=0.66 ] (410.19,56.8) -- (530.19,56.8) -- (530.19,86.37) -- (410.19,86.37) -- cycle ;
\draw  [draw opacity=0][fill={rgb, 255:red, 80; green, 227; blue, 194 }  ,fill opacity=0.66 ] (272.97,55.8) -- (394.86,55.8) -- (394.86,85.37) -- (272.97,85.37) -- cycle ;
\draw  [draw opacity=0][fill={rgb, 255:red, 153; green, 212; blue, 255 }  ,fill opacity=0.68 ] (272.97,169) -- (530.19,169) -- (530.19,209.7) -- (272.97,209.7) -- cycle ;
\draw  [draw opacity=0][fill={rgb, 255:red, 255; green, 244; blue, 121 }  ,fill opacity=0.77 ] (409.52,20.73) -- (529.86,20.73) -- (529.86,50.3) -- (409.52,50.3) -- cycle ;
\draw    (402.5,6.23) -- (402.5,39.67) -- (402.5,81.95) ;
\draw (364.72,31.55) node  {\includegraphics[width=34.13pt,height=34.13pt]{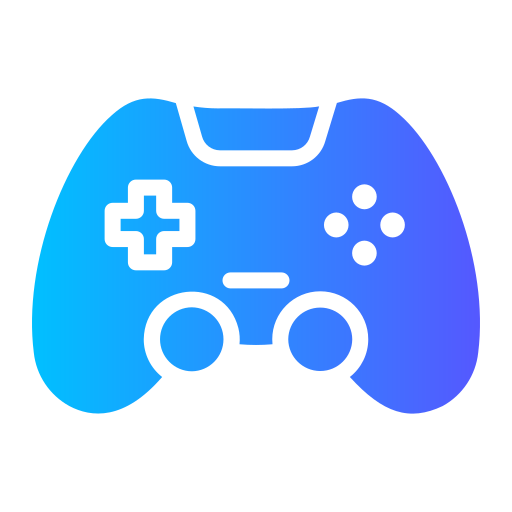}};
\draw (458.89,223.32) node  {\includegraphics[width=90.14pt,height=90.14pt]{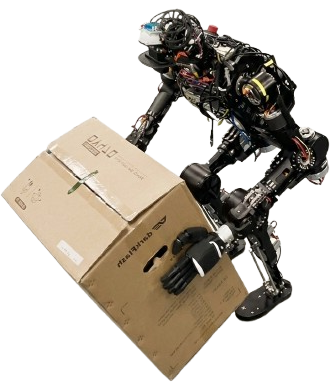}};

\draw (401.91,104.15) node   [align=left] {Centroidal MPC \ };
\draw (322.42,143.36) node   [align=left] {WBC};
\draw (333.91,70.58) node   [align=left] {Trajectory Planner};
\draw (468.35,35.52) node   [align=left] {\begin{minipage}[lt]{92.88pt}\setlength\topsep{0pt}
\begin{center}
Desired height/pitch
\end{center}

\end{minipage}};
\draw (280.5,25.2) node [anchor=north west][inner sep=0.75pt]   [align=left] {walking};
\draw (412.5,1.7) node [anchor=north west][inner sep=0.75pt]   [align=left] {posture transition};
\draw (454.36,142.61) node  [font=\footnotesize] [align=left] {\begin{minipage}[lt]{121.11pt}\setlength\topsep{0pt}
\begin{center}
Adaptive Fuzzy Balance Control \\with Torque Allocation 
\end{center}

\end{minipage}};
\draw (469.85,71.58) node   [align=left] {\begin{minipage}[lt]{74.74pt}\setlength\topsep{0pt}
\begin{center}
Motion planning
\end{center}

\end{minipage}};
\draw (269.97,117.44) node [anchor=south west] [inner sep=0.75pt]  [rotate=-270] [align=left] {50Hz};
\draw (269.97,159.79) node [anchor=south west] [inner sep=0.75pt]  [rotate=-270] [align=left] {500Hz};
\draw (269.97,207.7) node [anchor=south west] [inner sep=0.75pt]  [rotate=-270] [align=left] {1kHz};
\draw (350,189.62) node  [font=\footnotesize] [align=left] {\begin{minipage}[lt]{92.53pt}\setlength\topsep{0pt}
\begin{center}
Humanoid robot platform\\+\\Low-level control
\end{center}

\end{minipage}};
\draw (342,241.62) node  [font=\normalsize] [align=left] {\begin{minipage}[lt]{84.94pt}\setlength\topsep{0pt}
\begin{center}
Unknown Payload
\end{center}

\end{minipage}};

\end{tikzpicture}

%% file: content/3_1_posture_transition.tex
\section{Optimal Posture-transition Motion Planning}

This section focuses on generating motions for posture transitions of the robot while standing, such as standing up, sitting down, and leaning backward. 

\begin{figure}
    \centering
    \scalebox{0.9}{\input{tikz/skeleton}}
    \caption{A humanoid robot model. The simplified skeleton of the robot is depicted on the right-hand side.}
    \label{fig:vrh3}
\end{figure}
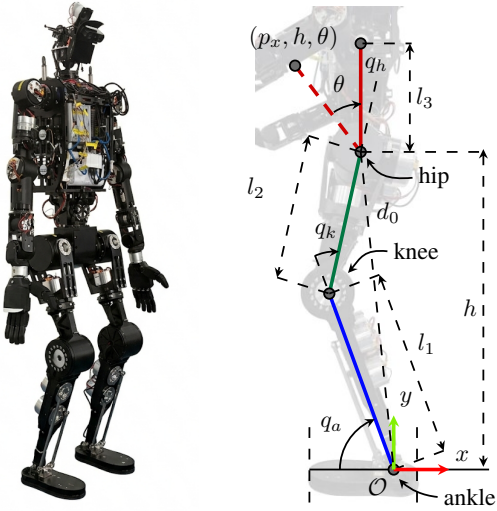

\subsection{Waypoint Model}
Consider the waypoint model of 3-link planar robot (shown in Fig.~\ref{fig:vrh3}) with the generalized coordinate $\mathbf{q}_w = \begin{bmatrix} q_{h} & q_{k} & q_{a} \end{bmatrix}^\top \in \mathbb{R}^{3}$, representing the hip, knee, and ankle joint angles. A smooth trajectory is generated using the discrete-time triple-integrator model
\begin{align}
\boldsymbol{\eta}_{k+1} = \boldsymbol{\Phi}\boldsymbol{\eta}_k + \boldsymbol{\Gamma}_1\boldsymbol{\omega}_k + \boldsymbol{\Gamma}_2\boldsymbol{\omega}_{k+1},
\end{align}
where $\boldsymbol{\eta} = \begin{bmatrix} \mathbf{q}_w^\top & \dot{\mathbf{q}}_w^\top & \ddot{\mathbf{q}}_w^\top \end{bmatrix}^\top$ denotes the joint positions, velocities, and accelerations, $\boldsymbol{\omega} = \dddot{\mathbf{q}}_w$ is the jerk input, and
{
\footnotesize
\begin{align}
\boldsymbol{\Phi} = \rev{\begin{bmatrix} 1 & \Delta t & \frac{\Delta t^2}{2} \\ 0 & 1 & \Delta t \\ 0 & 0 & 1 \end{bmatrix}}\otimes \mathbf{I}_3,\quad
\boldsymbol{\Gamma}_1 = \rev{\begin{bmatrix} \frac{\Delta t^3}{8} \\ \frac{\Delta t^2}{3} \\ \frac{\Delta t}{2} \end{bmatrix}}\otimes \mathbf{I}_3,\quad
\boldsymbol{\Gamma}_2 = \rev{\begin{bmatrix} \frac{\Delta t^3}{24} \\ \frac{\Delta t^2}{6} \\ \frac{\Delta t}{2} \end{bmatrix}}\otimes \mathbf{I}_3, \notag
\end{align}
}where $\otimes$ denotes the Kronecker product.

\subsection{Pelvis posture selection mechanism}
In this case, the analytical inverse kinematics is relatively straightforward. However, selecting the suitable position of the pelvis under the commands of pelvis height and pelvis pitch motion is problematic. With the desired pelvis height and pitch rotation, there is an infinite number of solutions for pelvis horizontal motion, such that the projection within the support polygon. Therefore, the Algorithm~\ref{alg:pelvis} shows how the horizontal position of the pelvis is selected.

\begin{algorithm}
\caption{Pelvis posture selection mechanism.}
\label{alg:pelvis}
\begin{algorithmic}[1]
\Require Desired pelvis height $h$, pitch rotation $\theta$, and resolution value $\varepsilon$
\State Initial $p_x \in \texttt{support\_polygon}$
\While{$p_x \in \texttt{support\_polygon}$}
    \State $\mathbf{q}_{wr} \gets \texttt{inverse\_kinematics}(h, p_x, \theta)$
    \If{$\mathbf{q}_{wr} \notin \mathcal{Q}$} \Comment{within joint limits}
        \State $p_x \gets p_x + \mathrm{sign}(\theta)\varepsilon$
    \Else
        \State \Return $\mathbf{q}_{wr}$
    \EndIf
\EndWhile
\State \Return No feasible joint configuration
\end{algorithmic}
\end{algorithm}

Regarding inverse kinematics, it is straightforward to derive via the law of cosines,
\begin{align}
q_k &= \pi - \arccos\!\left(\frac{l_1^2 + l_2^2 - d_0^2}{2 l_1 l_2} \right),\\
q_h &= \frac{\pi}{2} - \arccos\!\left(\frac{l_1^2 + d_0^2 - l_2^2}{2 l_1 d_0}\right) - \arctan2(h, p_x), \\
q_a &= -q_h - q_k - \theta,
\end{align}
where $l_0$, $l_1$, and $l_2$ are the link length of 3-link planar robot, while $d_0$ is the distance from the hip to ankle joint. The sign of $\arccos$ could be easily selected by the constraint in the robot joint limitation. 

\begin{figure}
    \centering
    \includegraphics[width=1.03\linewidth]{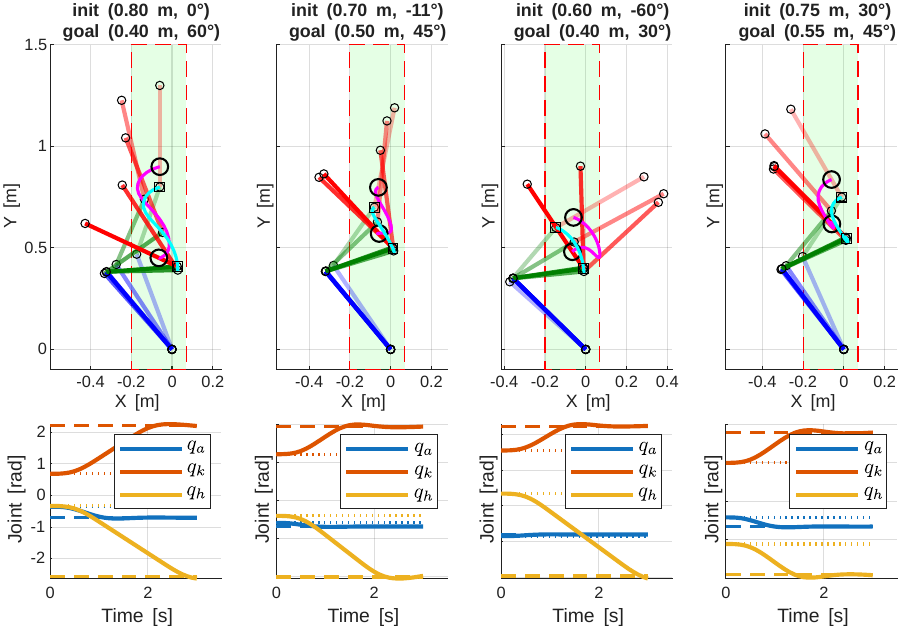}
    \vspace{0.01cm}
    \caption{Examples of humanoid robot posture transition. The rectangular marker indicates the initial and goal posture of the pelvis, while the circular marker indicates the initial and goal posture of CoM. Furthermore, its trajectory during the transition is shown by cyan and purple curves. The green transparent ranges indicate the boundary constraints of CoM}
    \label{fig:posture transition}
\end{figure}a

\subsection{Posture-transition motion generation}

\begin{figure*}
    \centering
    \includegraphics[width=1\linewidth]{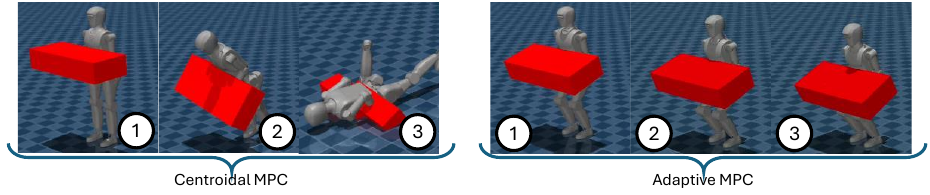}
    \caption{Simulation of holding an unknown heavy payload. (Left) The standard Centroidal MPC without posture planning fails to compensate for the added mass, causing the robot to tip forward and collapse. (Right) The proposed Adaptive MPC with posture Transition Generation dynamically lowers the posture to counteract the disturbance, successfully maintaining balance while holding the payload.}
    \label{fig:sim}
\end{figure*}

Given the desired pelvis posture $(h,\theta)$, the reference joint configuration $\mathbf{q}_{wr}$ is obtained from the inverse kinematics. The motion planning problem is then formulated as the following constrained optimization problem
\begin{subequations}\label{eq:mpc}
\begin{align}
\min_{\boldsymbol{\eta}, \boldsymbol{\omega}} 
& \quad \sum_{k=0}^{N} 
\left(
\|\rev{\mathbf{q}_{w,k}}-\mathbf{q}_{wr, k}\|_{\mathbf{W}_x}^2 
+ 
\|\boldsymbol{\omega}_k\|_{\mathbf{W}_u}^2
\right)
\label{eq:mpc_obj} \\[2mm]
\text{s.t.}\quad
& \boldsymbol{\eta}_{k+1}  = 
\boldsymbol{\Phi}\boldsymbol{\eta}_k 
+ 
\boldsymbol{\Gamma}_1\boldsymbol{\omega}_k 
+ 
\boldsymbol{\Gamma}_2\boldsymbol{\omega}_{k+1}
\label{eq:mpc_dyn} \\
&\mathbf{q}_{w0} = \mathbf{q}_{w, \mathrm{init}} \label{eq:init} \\
& \boldsymbol{\eta}_{k} \in \mathcal{N},
\quad \boldsymbol{\omega}_{k} \in \mathcal{W} \label{eq:mpc_state}  \\
& \gamma(\rev{\mathbf{q}_{w,k}}) \in \texttt{support\_polygon}\label{eq:mpc_com}
\end{align}
\end{subequations}
In \eqref{eq:mpc}, \eqref{eq:mpc_obj} defines the quadratic cost function of the tracking error w.r.t. the reference configuration $\mathbf{q}_r$ and the jerk input $\boldsymbol{\omega}$, weighted by $\mathbf{W}_x$ and $\mathbf{W}_u$, respectively. Constraint \eqref{eq:mpc_dyn} represents the discrete-time triple-integrator dynamics with the initial states constraint \eqref{eq:init}. Constraints \eqref{eq:mpc_state} impose bounds on joint positions, velocities, accelerations, and jerk, within the set $\mathcal{N}$ of $\boldsymbol{\eta}$ and set $\mathcal{W}$ of $\boldsymbol{\omega}$. Finally, constraint \eqref{eq:mpc_com} ensures that the Center of Mass, represented by $\gamma(\mathbf{q}_k)$, remains within the support polygon to guarantee balance.

The Fig.~\ref{fig:posture transition} illustrates the pose transition for the robot under the proposed method. As shown in the results, the transition is smooth, and in all scenarios, the center of mass projection is within the support polygon. Besides that, the $x$-direction of the initial and final positions of the pelvis is changed. This is due to the change in $p_x$ in Algorithm~\ref{alg:pelvis}, which helps the robot find a feasible pose with the desired height and pitch.

%% file: tikz/skeleton.tex
\tikzset{every picture/.style={line width=0.75pt}} 

\begin{tikzpicture}[x=0.75pt,y=0.75pt,yscale=-1,xscale=1]

\draw (489,161.55) node  {\includegraphics[width=102pt,height=223.57pt]{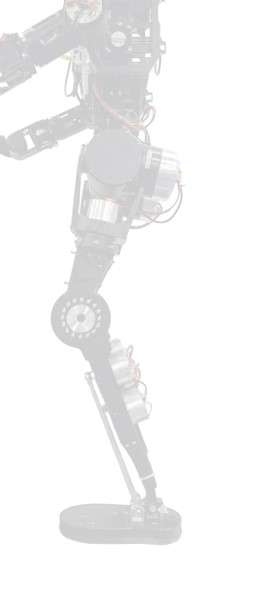}};
\draw    (551.5,270.9) -- (498.3,270.9) ;
\draw [color={rgb, 255:red, 206; green, 0; blue, 0 }  ,draw opacity=1 ][line width=1.5]  [dash pattern={on 5.63pt off 4.5pt}]  (443.3,45.7) -- (479.9,93.8) ;
\draw    (451.3,270.9) -- (511.8,270.9) ;
\draw [color={rgb, 255:red, 0; green, 6; blue, 255 }  ,draw opacity=1 ][line width=1.5]    (498.3,270.9) -- (462.4,172.8) ;
\draw [color={rgb, 255:red, 1; green, 119; blue, 67 }  ,draw opacity=1 ][line width=1.5]    (479.9,93.8) -- (462.4,172.8) ;
\draw [color={rgb, 255:red, 206; green, 0; blue, 0 }  ,draw opacity=1 ][line width=1.5]    (479.9,33.3) -- (479.9,93.8) ;
\draw  [fill={rgb, 255:red, 105; green, 105; blue, 105 }  ,fill opacity=1 ] (477,93.8) .. controls (477,92.2) and (478.3,90.9) .. (479.9,90.9) .. controls (481.5,90.9) and (482.8,92.2) .. (482.8,93.8) .. controls (482.8,95.4) and (481.5,96.7) .. (479.9,96.7) .. controls (478.3,96.7) and (477,95.4) .. (477,93.8) -- cycle ;
\draw  [fill={rgb, 255:red, 105; green, 105; blue, 105 }  ,fill opacity=1 ] (477,33.3) .. controls (477,31.7) and (478.3,30.4) .. (479.9,30.4) .. controls (481.5,30.4) and (482.8,31.7) .. (482.8,33.3) .. controls (482.8,34.9) and (481.5,36.2) .. (479.9,36.2) .. controls (478.3,36.2) and (477,34.9) .. (477,33.3) -- cycle ;
\draw  [fill={rgb, 255:red, 105; green, 105; blue, 105 }  ,fill opacity=1 ] (459.5,172.8) .. controls (459.5,171.2) and (460.8,169.9) .. (462.4,169.9) .. controls (464,169.9) and (465.3,171.2) .. (465.3,172.8) .. controls (465.3,174.4) and (464,175.7) .. (462.4,175.7) .. controls (460.8,175.7) and (459.5,174.4) .. (459.5,172.8) -- cycle ;
\draw  [fill={rgb, 255:red, 105; green, 105; blue, 105 }  ,fill opacity=1 ] (495.4,270.9) .. controls (495.4,269.3) and (496.7,268) .. (498.3,268) .. controls (499.9,268) and (501.2,269.3) .. (501.2,270.9) .. controls (501.2,272.5) and (499.9,273.8) .. (498.3,273.8) .. controls (496.7,273.8) and (495.4,272.5) .. (495.4,270.9) -- cycle ;
\draw  [dash pattern={on 4.5pt off 4.5pt}]  (462.4,172.8) -- (490.3,162.3) ;
\draw  [dash pattern={on 4.5pt off 4.5pt}]  (480.9,33.3) -- (508.3,33.3) ;
\draw  [dash pattern={on 4.5pt off 4.5pt}]  (491.33,165.12) -- (525.17,257.58) ;
\draw [shift={(526.2,260.4)}, rotate = 249.9] [fill={rgb, 255:red, 0; green, 0; blue, 0 }  ][line width=0.08]  [draw opacity=0] (5.36,-2.57) -- (0,0) -- (5.36,2.57) -- (3.56,0) -- cycle    ;
\draw [shift={(490.3,162.3)}, rotate = 69.9] [fill={rgb, 255:red, 0; green, 0; blue, 0 }  ][line width=0.08]  [draw opacity=0] (5.36,-2.57) -- (0,0) -- (5.36,2.57) -- (3.56,0) -- cycle    ;
\draw  [dash pattern={on 4.5pt off 4.5pt}]  (498.3,270.9) -- (526.2,260.4) ;
\draw  [dash pattern={on 4.5pt off 4.5pt}]  (433.3,163.8) -- (462.4,172.8) ;
\draw  [dash pattern={on 4.5pt off 4.5pt}]  (450.8,84.8) -- (479.9,93.8) ;
\draw  [dash pattern={on 4.5pt off 4.5pt}]  (450.15,87.73) -- (433.95,160.87) ;
\draw [shift={(433.3,163.8)}, rotate = 282.49] [fill={rgb, 255:red, 0; green, 0; blue, 0 }  ][line width=0.08]  [draw opacity=0] (5.36,-2.57) -- (0,0) -- (5.36,2.57) -- (3.56,0) -- cycle    ;
\draw [shift={(450.8,84.8)}, rotate = 102.49] [fill={rgb, 255:red, 0; green, 0; blue, 0 }  ][line width=0.08]  [draw opacity=0] (5.36,-2.57) -- (0,0) -- (5.36,2.57) -- (3.56,0) -- cycle    ;
\draw  [dash pattern={on 4.5pt off 4.5pt}]  (479.9,93.8) -- (548.3,93.8) ;
\draw  [dash pattern={on 4.5pt off 4.5pt}]  (507.3,36.3) -- (507.3,90.8) ;
\draw [shift={(507.3,93.8)}, rotate = 270] [fill={rgb, 255:red, 0; green, 0; blue, 0 }  ][line width=0.08]  [draw opacity=0] (5.36,-2.57) -- (0,0) -- (5.36,2.57) -- (3.56,0) -- cycle    ;
\draw [shift={(507.3,33.3)}, rotate = 90] [fill={rgb, 255:red, 0; green, 0; blue, 0 }  ][line width=0.08]  [draw opacity=0] (5.36,-2.57) -- (0,0) -- (5.36,2.57) -- (3.56,0) -- cycle    ;
\draw  [draw opacity=0] (468.3,270.84) .. controls (468.33,258.01) and (476.4,247.08) .. (487.74,242.81) -- (498.3,270.9) -- cycle ; \draw    (468.3,270.84) .. controls (468.32,259.04) and (475.16,248.84) .. (485.08,243.96) ; \draw [shift={(487.74,242.81)}, rotate = 153.82] [fill={rgb, 255:red, 0; green, 0; blue, 0 }  ][line width=0.08]  [draw opacity=0] (5.36,-2.57) -- (0,0) -- (5.36,2.57) -- (3.56,0) -- cycle    ; 
\draw  [dash pattern={on 4.5pt off 4.5pt}]  (454.49,151.73) -- (462.4,172.8) ;
\draw  [draw opacity=0] (454.19,151.87) .. controls (457.67,150.22) and (461.55,149.3) .. (465.65,149.3) .. controls (466.36,149.3) and (467.06,149.33) .. (467.76,149.38) -- (465.65,176.05) -- cycle ; \draw    (454.19,151.87) .. controls (457.67,150.22) and (461.55,149.3) .. (465.65,149.3) ; \draw [shift={(467.76,149.38)}, rotate = 178.08] [fill={rgb, 255:red, 0; green, 0; blue, 0 }  ][line width=0.08]  [draw opacity=0] (5.36,-2.57) -- (0,0) -- (5.36,2.57) -- (3.56,0) -- cycle    ; 
\draw  [dash pattern={on 4.5pt off 4.5pt}]  (488.65,54.3) -- (479.9,93.8) ;
\draw  [dash pattern={on 4.5pt off 4.5pt}]  (451.3,247.3) -- (451.3,291.8) ;
\draw  [dash pattern={on 4.5pt off 4.5pt}]  (511.3,247.3) -- (511.3,291.8) ;
\draw    (508.8,108.2) .. controls (494.12,109.53) and (488.94,106.12) .. (484.8,99.71) ;
\draw [shift={(483.3,97.2)}, rotate = 60.64] [fill={rgb, 255:red, 0; green, 0; blue, 0 }  ][line width=0.08]  [draw opacity=0] (5.36,-2.57) -- (0,0) -- (5.36,2.57) -- (3.56,0) -- cycle    ;
\draw    (494.8,149.2) .. controls (481.82,147.86) and (479.66,152.92) .. (474.69,159.09) ;
\draw [shift={(472.8,161.3)}, rotate = 312.47] [fill={rgb, 255:red, 0; green, 0; blue, 0 }  ][line width=0.08]  [draw opacity=0] (5.36,-2.57) -- (0,0) -- (5.36,2.57) -- (3.56,0) -- cycle    ;
\draw    (522.8,286.2) .. controls (514.25,285.34) and (507.89,282.3) .. (503.42,278.31) ;
\draw [shift={(501.3,276.2)}, rotate = 48.01] [fill={rgb, 255:red, 0; green, 0; blue, 0 }  ][line width=0.08]  [draw opacity=0] (5.36,-2.57) -- (0,0) -- (5.36,2.57) -- (3.56,0) -- cycle    ;
\draw [color={rgb, 255:red, 255; green, 0; blue, 0 }  ,draw opacity=1 ][line width=1.5]    (498.3,270.9) -- (525.3,270.9) ;
\draw [shift={(529.3,270.9)}, rotate = 180] [fill={rgb, 255:red, 255; green, 0; blue, 0 }  ,fill opacity=1 ][line width=0.08]  [draw opacity=0] (6.43,-3.09) -- (0,0) -- (6.43,3.09) -- (4.27,0) -- cycle    ;
\draw [color={rgb, 255:red, 131; green, 240; blue, 4 }  ,draw opacity=1 ][line width=1.5]    (498.3,270.9) -- (498.3,244.7) ;
\draw [shift={(498.3,240.7)}, rotate = 90] [fill={rgb, 255:red, 131; green, 240; blue, 4 }  ,fill opacity=1 ][line width=0.08]  [draw opacity=0] (6.43,-3.09) -- (0,0) -- (6.43,3.09) -- (4.27,0) -- cycle    ;
\draw  [dash pattern={on 4.5pt off 4.5pt}]  (548.3,95.2) -- (548.3,266.4) ;
\draw [shift={(548.3,269.4)}, rotate = 270] [fill={rgb, 255:red, 0; green, 0; blue, 0 }  ][line width=0.08]  [draw opacity=0] (5.36,-2.57) -- (0,0) -- (5.36,2.57) -- (3.56,0) -- cycle    ;
\draw [shift={(548.3,92.2)}, rotate = 90] [fill={rgb, 255:red, 0; green, 0; blue, 0 }  ][line width=0.08]  [draw opacity=0] (5.36,-2.57) -- (0,0) -- (5.36,2.57) -- (3.56,0) -- cycle    ;
\draw  [fill={rgb, 255:red, 105; green, 105; blue, 105 }  ,fill opacity=1 ] (440.4,45.7) .. controls (440.4,44.1) and (441.7,42.8) .. (443.3,42.8) .. controls (444.9,42.8) and (446.2,44.1) .. (446.2,45.7) .. controls (446.2,47.3) and (444.9,48.6) .. (443.3,48.6) .. controls (441.7,48.6) and (440.4,47.3) .. (440.4,45.7) -- cycle ;
\draw  [draw opacity=0] (464.93,71.63) .. controls (469.2,68.74) and (474.35,67.05) .. (479.9,67.05) .. controls (480.03,67.05) and (480.17,67.05) .. (480.3,67.05) -- (479.9,93.8) -- cycle ; \draw    (464.93,71.63) .. controls (469.2,68.74) and (474.35,67.05) .. (479.9,67.05) ; \draw [shift={(480.3,67.05)}, rotate = 174.48] [fill={rgb, 255:red, 0; green, 0; blue, 0 }  ][line width=0.08]  [draw opacity=0] (5.36,-2.57) -- (0,0) -- (5.36,2.57) -- (3.56,0) -- cycle    ; 
\draw  [dash pattern={on 4.5pt off 4.5pt}]  (479.9,93.8) -- (498.3,270.9) ;
\draw (321.24,149.5) node  {\includegraphics[height=222.75pt]{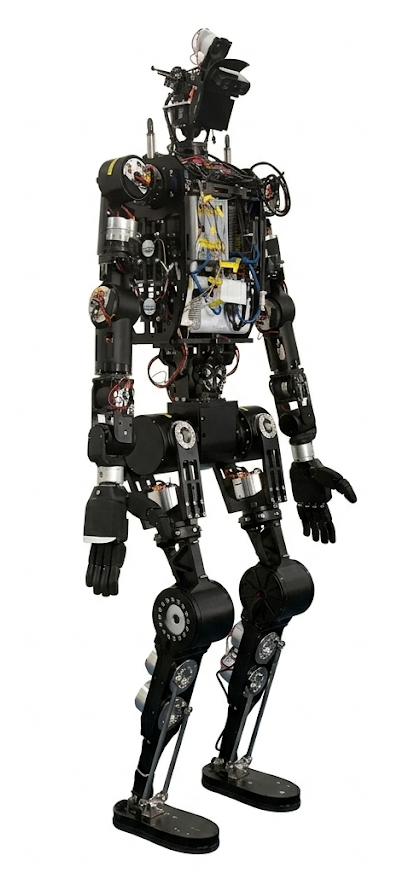}};

\draw (510.25,207.95) node [anchor=south west] [inner sep=0.75pt]    {$l_{1}$};
\draw (428.05,120.9) node [anchor=south east] [inner sep=0.75pt]    {$l_{2}$};
\draw (509.3,63.55) node [anchor=west] [inner sep=0.75pt]    {$l_{3}$};
\draw (471.3,251.44) node [anchor=south east] [inner sep=0.75pt]    {$q_{a}$};
\draw (460.49,143.83) node [anchor=south] [inner sep=0.75pt]    {$q_{k}$};
\draw (488.65,50.9) node [anchor=south] [inner sep=0.75pt]    {$q_{h}$};
\draw (510.8,108.2) node [anchor=west] [inner sep=0.75pt]   [align=left] {hip};
\draw (496.8,149.2) node [anchor=west] [inner sep=0.75pt]   [align=left] {knee};
\draw (524.8,286.2) node [anchor=west] [inner sep=0.75pt]   [align=left] {ankle};
\draw (496.3,274.3) node [anchor=north east] [inner sep=0.75pt]    {$\mathcal{O}$};
\draw (531.3,267.5) node [anchor=south west] [inner sep=0.75pt]    {$x$};
\draw (500.3,237.3) node [anchor=south west] [inner sep=0.75pt]    {$y$};
\draw (463,49.4) node [anchor=north west][inner sep=0.75pt]    {$\theta $};
\draw (546.3,180.8) node [anchor=east] [inner sep=0.75pt]    {$h$};
\draw (416,21.9) node [anchor=north west][inner sep=0.75pt]    {$( p_{x} ,h,\theta )$};
\draw (487.25,134.95) node [anchor=south west] [inner sep=0.75pt]    {$d_{0}$};

\end{tikzpicture}

%% file: content/4_experiment.tex

\begin{figure}
    \centering
    \includegraphics[width=0.95\linewidth]{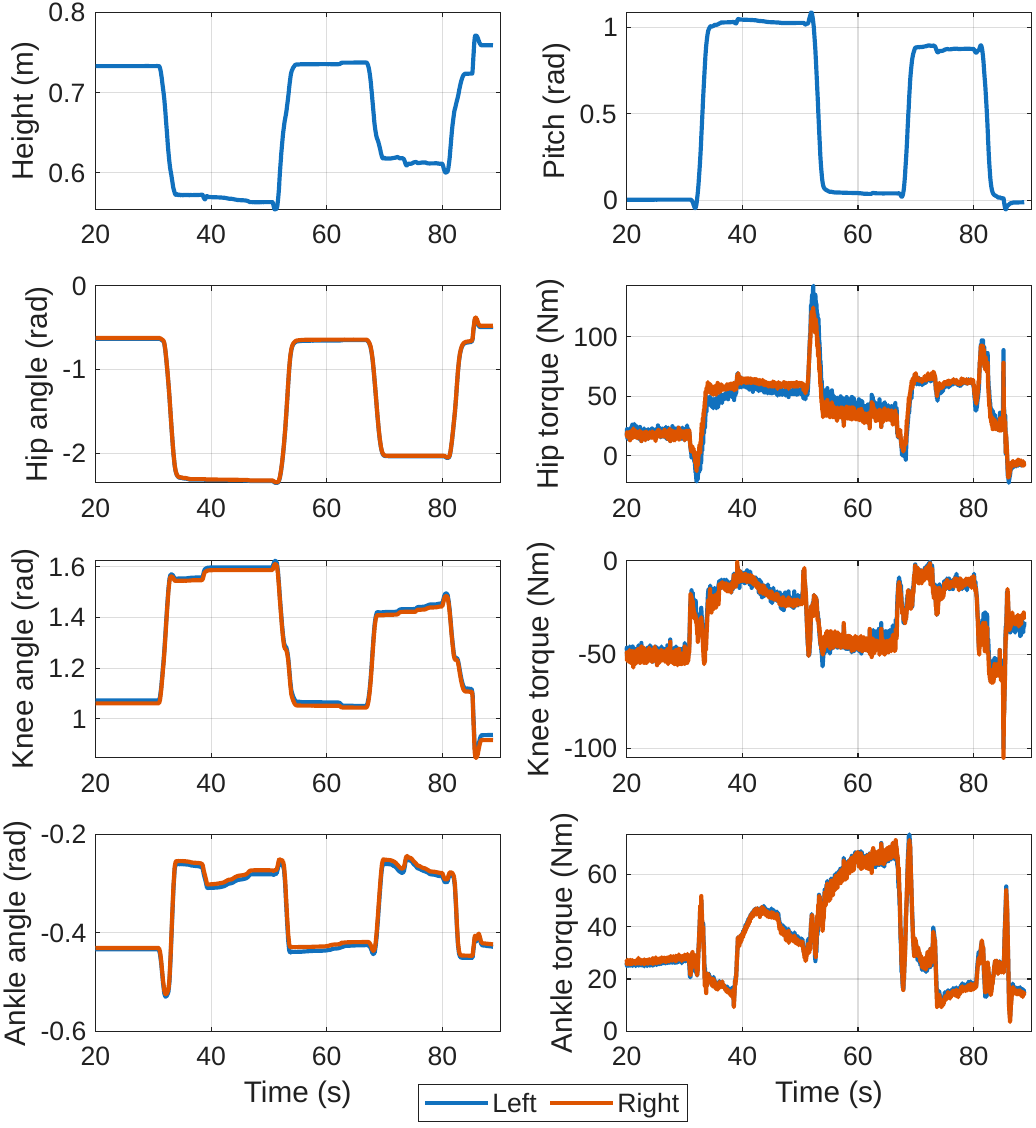}
    \vspace{0.1cm}
    \caption{Experimental results for the 7 kg payload lifting task.}
    \label{fig:7kg}
\end{figure}

\begin{figure}
    \centering
    \includegraphics[width=0.95\linewidth]{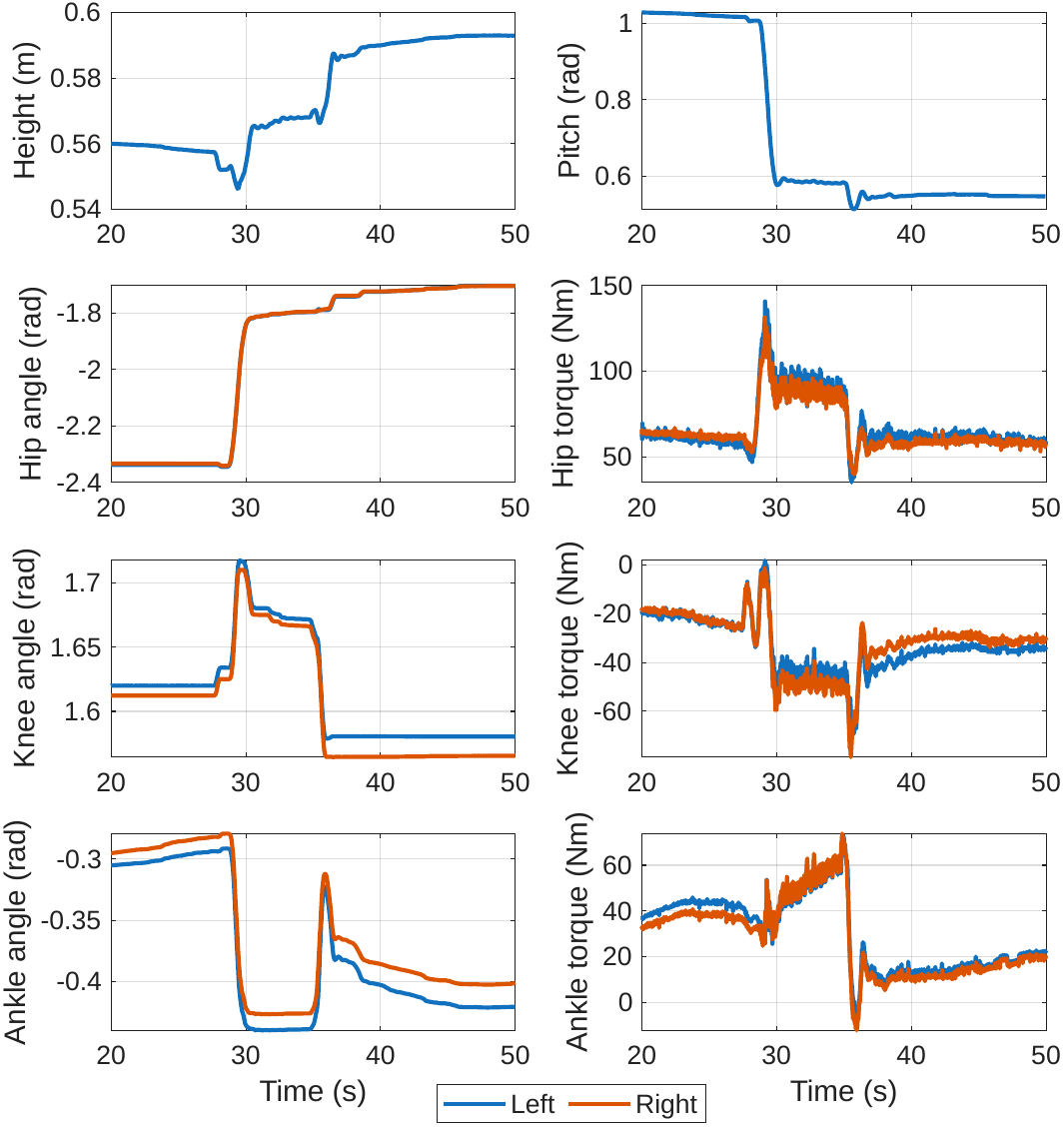}
    \vspace{0.1cm}
    \caption{Experimental results for the 10 kg payload lifting task.}
    \label{fig:10kg}
\end{figure}

\begin{figure}
    \centering
    \includegraphics[width=\linewidth]{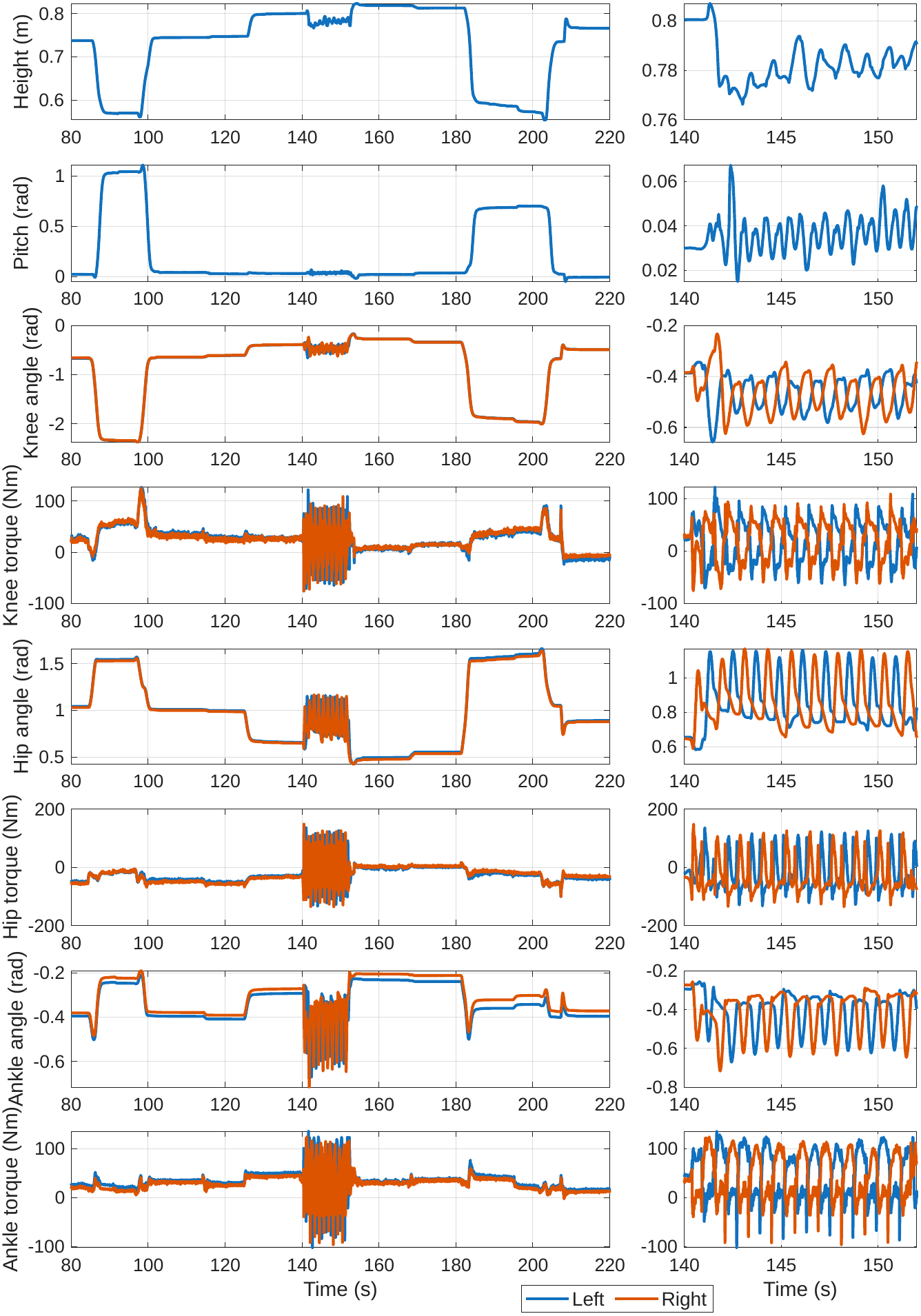}
    \vspace{0.01cm}
    \caption{Experimental results of lifting-walking scenario.}
    \label{fig:walking}
\end{figure}

\section{Experimental Results}
\label{sec: experiment}

In this section, simulations and experiments are conducted to demonstrate the performance of the proposed approach under ideal and real-world conditions. The proposed method is implemented on a Humanoid Robot shown in Fig.~\ref{fig:vrh3}, which has 28 DoFs (7 for each arm and 6 for each leg).

The overall control structure based on the proposed approach is illustrated in Fig.~\ref{fig:structure}. The framework supports two operating modes. The first mode enables manual control via a joystick, combined with a trajectory planner that generates reference velocities for the pelvis. The second mode focuses on posture transition planning, which is handled by the proposed optimization method described in Section IV.

The reference pelvis states are then provided to the centroidal model MPC, which operates at 50 Hz. Subsequently, the proposed adaptive control framework, running at 500 Hz, generates the desired joint torques. These torque commands are finally executed by a low-level controller operating at a frequency above 1 kHz, ensuring fast and stable torque tracking.

\subsection{Simulation validation}

Simulation validations were conducted within the MuJoCo physics engine \cite{todorov2012mujoco} to independently evaluate the effectiveness of the proposed modules. As illustrated in Fig.~\ref{fig:sim}, when lifting an unknown payload using a standard Centroidal MPC (left), the lack of an adaptive balancing control leads to uncompensated CoM shifts. This, compounded by non-optimized kinematic trajectories, causes the robot to exceed its stability margin and collapse (Stages 1–3). In contrast, the proposed framework (right) successfully maintains whole-body equilibrium. By integrating the Adaptive MPC with the Posture Transition Planner, the robot dynamically computes the required compensation torques and synthesizes feasible waypoints to proactively lower its center of mass (Stages 2–3). This synergy ensures that the CoM remains rigorously within the stability margin throughout the entire manipulation cycle, highlighting the necessity of combining adaptive disturbance rejection with task-optimal posture generation. \rev{This keeps the CoM within the stability margin throughout the manipulation cycle.}



\subsection{Experiment implementation}

Two experimental scenarios are conducted. First, a static lifting task is considered to evaluate the robot’s ability to handle payload variations. Second, a dynamic scenario combining lifting and load-carrying during locomotion is implemented to further assess the performance and robustness of the proposed control strategy.

\subsubsection{Lifting}

In the lifting experiment scenario, two payloads of 7 kg and 10 kg were implemented. 

Figure~\ref{fig:7kg} presents the experimental results for the 7 kg payload case. The results include the evolution of the pelvis height and pitch angle, as well as the rotation and torque profiles of the hip, knee, and ankle joints. With the proposed motion planning strategy, smooth transitions in posture and joint states are achieved throughout the task. At approximately 30s, the robot begins to decrease its pelvis height and lean forward in pitch, completing the sitting motion within 3s. During the subsequent phase, the robot grasps the 7 kg object and initiates the lifting process. After successfully lifting the object, the robot stands up during the period from 55s to 80s before finally placing the object back down.

For the 10 kg payload, the experimental results are presented in Fig.~\ref{fig:10kg}. In this scenario, the object mass is significantly increased, which requires greater control effort during the lifting task. At approximately 20 s, the robot moves into a bowing posture in preparation for lifting. After grasping the object at around 30 s, the robot slightly lifts it while gradually recovering the torso pitch angle.

Due to the substantial change in payload mass and the resulting variation in the torso dynamics, the adaptive controller increases its control effort to maintain balance. This behavior is reflected by small oscillations in the pelvis height and pitch angle during the initial lifting phase. Once the pitch angle is stabilized, the robot continues to recover its height and eventually returns to the nominal upright position.

\subsubsection{Lifting and Walking}

In the combined lifting and walking scenario, a structured task pipeline is implemented. \rev{The carried object weighs 7\,kg.} The experimental results are presented in Fig.~\ref{fig:punch} and Fig.~\ref{fig:walking}. The left panel illustrates the evolution of the robot states over time, while the right panel provides a zoomed view of the walking phase. The proposed pipeline consists of the following stages: 1) The robot starts from a stable home position, then bows down and grasps the object (83s); 2) After grasping, the robot lifts the object and returns to the home posture (110s); 3) The robot walks while carrying the load for 15s, from 140s to 155s, before stopping; 4) Finally, the robot places the object on the ground and recovers to the home position. The results demonstrate that, under uncertain payload conditions, the robot is able to walk normally and maintain stable locomotion while successfully completing the assigned task. \rev{The fuzzy parameters were not re-tuned for walking, indicating that the tuning is not lifting-specific.}

%% file: content/5_discussion.tex
\section{Conclusion}



This paper presented an adaptive whole-body control architecture for humanoid robots handling unknown payloads by integrating Optimal Posture Planning, Whole-Body MPC, and Adaptive Fuzzy Control. The framework proactively generates dynamically feasible motions while the fuzzy controller reactively compensates for payload-induced torque disturbances, preserving the optimal and constraint-consistent behavior of the MPC. Validations in MuJoCo and on a real-world platform across stationary lifting and dynamic loco-manipulation scenarios demonstrated the system's effectiveness in maintaining whole-body balance. To overcome the heuristic and reactive limitations of the current fuzzy logic, future work will focus on integrating learning-based methods to optimize compensation strategies and proactive disturbance observers to further enhance agility in highly dynamic environments.